\documentclass[letterpaper, 10 pt, conference]{ieeeconf}
\IEEEoverridecommandlockouts
\let\labelindent\relax
\let\labelindent\relax
\usepackage{graphicx} \graphicspath{ {figures/} }
\usepackage{amsmath,amssymb,mathabx,amsbsy,mathtools,etoolbox}
\usepackage[utf8]{inputenc} 
\usepackage[T1]{fontenc}    

\usepackage{algorithmic}
\usepackage[ruled]{algorithm2e}

\usepackage{acronym}
\usepackage{enumitem}
\usepackage{balance}
\usepackage{xspace}
\usepackage{setspace}
\usepackage[skip=3pt,font=small]{subcaption}
\usepackage[skip=3pt,font=small]{caption}
\usepackage[dvipsnames,svgnames,x11names]{xcolor}
\usepackage[capitalise,nameinlink]{cleveref}
\usepackage{booktabs,tabularx,colortbl,multirow,multicol,array,makecell}
\usepackage{pifont}
\usepackage{cite}
\usepackage{nicematrix}

\makeatletter
\DeclareRobustCommand\onedot{\futurelet\@let@token\@onedot}
\def\@onedot{\ifx\@let@token.\else.\null\fi\xspace}

\def\etal{\emph{et al}\onedot}
\makeatother

\makeatletter
\def\BState{\State\hskip-\ALG@thistlm}
\makeatother

\makeatletter
\renewcommand{\paragraph}{%
  \@startsection{paragraph}{4}%
  {\z@}{0ex \@plus 0ex \@minus 0ex}{-1em}%
  {\hskip\parindent\normalfont\normalsize\bfseries}%
}
\makeatother

\crefname{algorithm}{Alg.}{Algs.}
\Crefname{algocf}{Algorithm}{Algorithms}
\crefname{section}{Sec.}{Secs.}
\Crefname{section}{Section}{Sections}
\crefname{table}{Tab.}{Tabs.}
\Crefname{table}{Table}{Tables}
\crefname{figure}{Fig.}{Fig.}
\Crefname{figure}{Figure}{Figure}

\definecolor{gblue}{HTML}{4285F4}
\definecolor{gred}{HTML}{DB4437}
\definecolor{ggreen}{HTML}{0F9D58}

\definecolor{mygray}{gray}{.92}

\acrodef{qp}[QP]{Quadratic Programming}
\acrodef{dof}[DoF]{Degree of Freedom}
\acrodef{ros}[ROS]{Robot Operating System}
\acrodef{uav}[UAV]{Unmanned Aerial Vehicle}
\acrodef{dof}[DoF]{Degree of Freedom}
\acrodef{com}[CoM]{center of mass}
\acrodef{msrr}[MSRR]{modular self-reconfigurable robot}
\acrodef{owmr}[OWMR]{Omnidirectional Wheeled Mobile Robot}
\acrodef{wmr}[WMR]{Wheeled Mobile Robot}

\title{\Large \bf Aggregating Single-wheeled Mobile Robots for Omnidirectional Movements}
\author{Meng Wang$^{1*}$\quad{}Yao Su$^{1*}$\quad{}Hang Li$^{1}$\quad{}Jiarui Li$^{1,2}$\quad{}Jixiang Liang$^{1,3}$\quad{}Hangxin Liu$^{1,\dagger}$%
\thanks{$^{*}$ Meng Wang and Yao Su contributed equally to this work. $\dagger$ Corresponding author. 
$^{1}$ National Key Laboratory of General Artificial Intelligence, Beijing Institute for General Artificial Intelligence (BIGAI).
$^{2}$  Department of Advanced Manufacturing and Robotics, College of Engineering, Peking University.
$^{3}$  Department of Robot Engineering, School of Automation Science and Electrical Engineering, Beihang University. Emails: \tt{\{wangmeng, suyao, lihang, lijiarui, liangjixiang, liuhx\}@bigai.ai.}}%
}

\begin{document}

\maketitle
\begin{abstract}
This paper presents a novel modular robot system that can self-reconfigure to achieve omnidirectional movements for collaborative object transportation. Each robotic module is equipped with a steerable omni-wheel for navigation and is shaped as a regular icositetragon with a permanent magnet installed on each corner for stable docking. After aggregating multiple modules and forming a structure that can cage a target object, we have developed an optimization-based method to compute the distribution of all wheels' heading directions, which enables efficient omnidirectional movements of the structure. By implementing a hierarchical controller on our prototyped system in both simulation and experiment, we validated the trajectory tracking performance of an individual module and a team of six modules in multiple navigation and collaborative object transportation settings. The results demonstrate that the proposed system can maintain a stable caging formation and achieve smooth transportation, indicating the effectiveness of our hardware and locomotion designs. 
\end{abstract}
\setstretch{0.98}
\section{Introduction}

Collaborative object transportation is a signature task that indicates how efficiently a team of robots can work together to accomplish what is beyond an individual's capability~\cite{tuci2018cooperative,alkilabi2017cooperative,an2023multi}. Indeed, being able to collaborate is not only a critical function in many animal societies and even in humans but also an intriguing approach to extend robots' capability in more challenging environments, such as in space~\cite{farivarnejad2021fully} and deep sea~\cite{pi2021twinbot}. Together with well-designed structures, communication mechanisms, and motion control algorithms, a number of swarm or modular robotic platforms have been developed~\cite{wang2020walkingbot,yim2000polybot,kim2016mroberto,eshaghi2020mroberto,yi2021puzzlebots,nisser2022electrovoxel}.  

The strategies for collaboratively transporting objects along with the reference trajectory can be roughly categorized into three types. In the first type, the \textit{pushing strategy}, several robots would actuate the object by generating contact force in certain directions~\cite{kube2000cooperative,chen2013strategy,chen2015occlusion}. 
Due to the lack of stable connections, the robots may lose control of the object during transportation, which leads to inefficient movements.
To tackle this issue, the \textit{grasping strategy} is developed as the second type. By installing connectors or manipulation mechanisms on each robot, they can build rigid connections with the object to ensure stable transportation and to better reject disturbance~\cite{machado2016multi,wang2016kinematic,hichri2016cooperative}. However, the extra mechanisms would dramatically increase the size and complexity of each module, reducing its adaptability in different transportation tasks. On the other hand, the \textit{caging strategy} for object transportation avoids the drawbacks and combines the benefits of the above two approaches. The team of robots would encircle the object through a structure formation to passively constrain or control the movement of the object~\cite{pereira2004decentralized,eoh2014distributed,wan2017multirobot,dai2016symmetric}. It does not require extra mechanisms to establish connections with the object, but it can stably push the caged object to a destination by navigating in a coordinated manner.

\begin{figure}[t!]
	\centering
 	\includegraphics[width=1\linewidth,trim=3cm 2cm 3cm 3cm, clip]{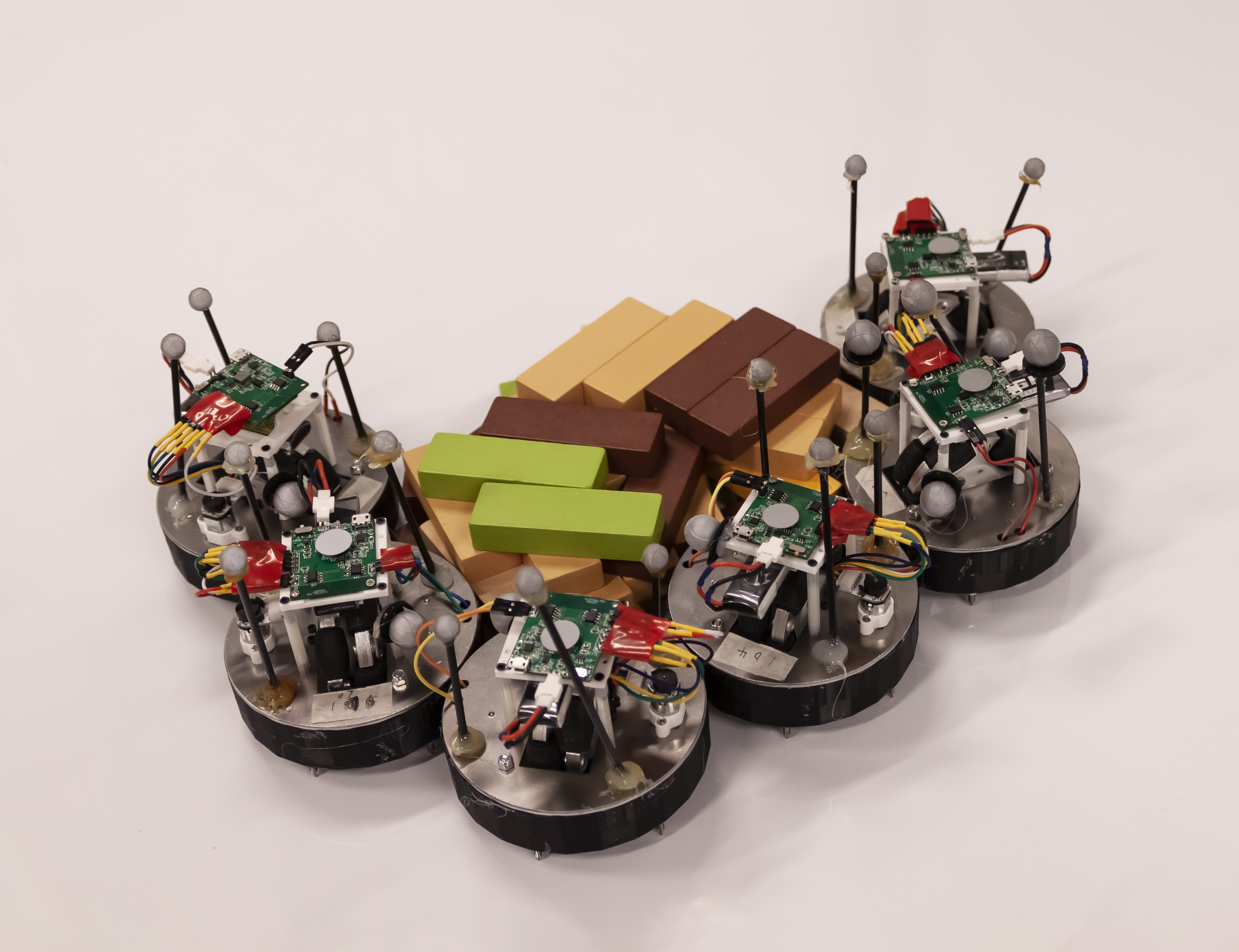}
 	\caption{\textbf{The robots are collaboratively transporting an object with complex shape through caging.} By magnetic docking with each other, robots can better maintain team formation to exert enough caging force while achieving omnidirectional movements for smooth transportation.}
	\label{fig:f1}
\end{figure}

However, implementing the caging strategy for a robot team is challenging. Firstly, the caging formation must be properly maintained throughout the transportation. Without rigid connections with each other or with the object, the caging forces exerted by the formation can easily deviate from desired values due to disturbance or possible collisions~\cite{wang2005algorithm}, especially when the caged object has a complex shape. In addition, maintaining structure formation when tracking trajectories with large curvature is very difficult, because each robot has to change its heading direction swiftly, and the team's movements must be well coordinated.

In this paper, we develop a novel modular robot system for efficient collaborative object transportation through caging. Each robot module in the team utilizes an omni-wheel with a steering mechanism for navigation. To strengthen the formation of the team, the contour of the robot is shaped into a regular icositetragon (twenty-four-sided polygon) with a permanent magnet installed on each corner for docking. Compared with a circular contour that commonly appeared in many other platforms~\cite{swissler2020fireant3d, tu2022freesn, zhao2022snailbot}, the polygon-based contour design provides extra shear strength to maintain structure formation while being flexible to establish many docking configurations. Therefore, the proposed robot system can encircle the target object with a complex shape and cage it robustly; see \cref{fig:f1}. Furthermore, by taking advantage of the steerable omni-wheel equipped by the robot, the structure can acquire an omnidirectional moving capability to transport the caged object more smoothly. An optimization framework is proposed to compute the heading direction of each robot's wheel such that the structure's controllability and energy efficiency are maximized for omnidirectional movements. 

Combining with a hierarchical controller, we verified the proposed robot system for collaborative object transportation in both simulation and real-world experiments. In the simulation, we evaluated whether the proposed optimization framework can produce the best heading direction of each robot's wheel to achieve the most efficient omnidirectional movements in collaborative object transportation. In experiments, four testing cases are studied to demonstrate (i) the locomotion of a single robot and of a structure consisting of six robots, (ii) the ability of the system to transport objects along a challenging trajectory, and (iii) the robustness of the hardware and controller design to transported object with heavy weight.

\subsection{Related Work}\label{sec:related}
In modular reconfigurable robots, various \textbf{docking mechanisms} were designed to strengthen the formed structure. Classical discrete docking designs~\cite{murata2002m,murata2006docking, romanishin20153d,neubert2016soldercubes,wei2010sambot} greatly constrained the number of possible assemble configurations. Therefore, continuous docking designs were further proposed with advanced scalability to meet more functional requirements, and it could be divided into 2D~\cite{li2019particle, shimizu2009amoeboid, kirby2007modular} and 3D~\cite{swissler2020fireant3d, tu2022freesn, zhao2022snailbot} branches depending on the type of connection. In particular, Li \etal proposed a passive-adhesion mechanism in~\cite{li2019particle}, the genderless Velcro strap and the Electromagnets ring were proposed in~\cite{shimizu2009amoeboid, kirby2007modular}, the permanent magnet approach was utilized in~\cite{tu2022freesn, zhao2022snailbot}, and Swissler \etal presented the self-soldering strategy~\cite{swissler2020fireant3d}. In this work, we utilize paired permanent magnets to approximate a continuous connector for our robot, which significantly enriches the caging configurations to adapt the geometry of different objects.

Changing routes and stabilizing carried objects through omnidirectional movements are more efficient during collaborative object transportation. Although equipping different types of wheels, such as the orthogonal wheel, spherical/ball wheel, and mecanum wheel~\cite{taheri2020omnidirectional,shen2020omburo}, can all achieve omnidirectional movements, they could be too bulky to build in small-scale mobile robots. 
Typical \ac{owmr} are designed with specific \textbf{transportation capacity} with four fixed omnidirectional wheels, resulting in poor scalability. To solve this issue, Almasri \etal proposed a novel \ac{owmr} design with multi symmetrically distributed omnidirectional wheels~\cite{almasri2021modeling} while 6-wheeled and 8-wheeled variants are presented in \cite{cao2022fuzzy,tian2017research}. In this work, we propose a novel modular design to effectively build 2D connected structures serving an \ac{owmr} system where each module is utilized as a steerable omni-directional wheel. Consequently, the transportation capacity of this system can be increased with more modules, and the \textbf{dynamic property} can also be modified with different topology structures or wheel heading angles.

\subsection{Overview}
We organize the remainder of the paper as follows. 
\cref{sec:design} presents the hardware design of our single-wheeled mobile robot.
\cref{sec:model} and \cref{sec:control} describe the dynamics and control of each
module and docked structure, respectively. \cref{sec:sim} and \cref{sec:exp} show the simulation and experiment results of our proposed system with comprehensive evaluations. Finally, we conclude the paper in \cref{sec:con}.

\section{Hardware Design}\label{sec:design}
This section describes the hardware design of the proposed modular robot system.
Each robot module in the system equips with one omni-wheel that can change its heading direction by a decoupled steering mechanism for self-navigation. A passive and continuous docking mechanism is developed to support flexible structure forming while maintaining a stable connection between robots.

\subsection{Integrated 2-DoF Driving Unit}
\begin{figure}[t!]
	\centering
 	\includegraphics[width=1\linewidth]{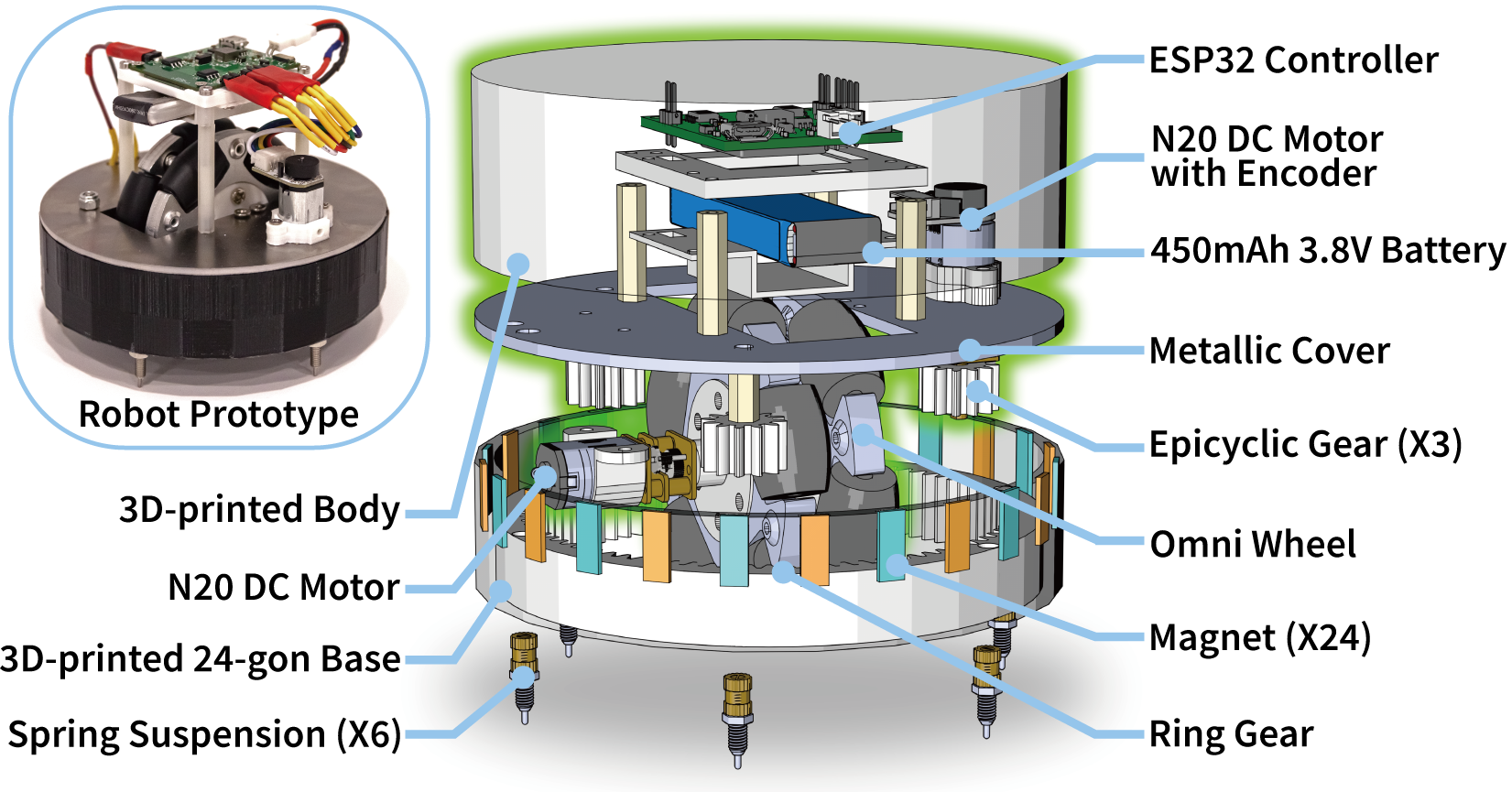}
 	\caption{\textbf{The hardware design of the robot.} The driving hat (green) integrates the  capability of steering and advancing through two DC motors. The omni-wheel supports omnidirectional movements after aggregating to a formation. The spring suspension can be adjusted to provide appropriate friction and keep the balance of the robot.  }
	\label{fig:design}
 \end{figure}

We propose an integrated 2-\ac{dof} robot unit design (see \cref{fig:design}) to facilitate flexible self-organization, self-formation, and self-reconfiguration. The robot consists of two parts: a docking base and a driving hat. Magnets are installed in the base, and spring-loaded ball rollers are placed at the bottom of the base serving as suspension supports. All electronic components are integrated into the hat, including a battery, a PCB, a DC motor driving an omni-wheel,  and a DC motor with an encoder for steering. The hat connects to the base through a set of planet/ring gears, such that the hat can  drive the planet gear to steer without changing the docking position. The omni-wheel in the center can  drive the robot forward along the steering direction. The bottom of the hat is manufactured with a slice of iron, which automatically attaches to the magnetic top of the base, and increases the friction of the driving wheel for better transportation performance.  
 
\begin{table}[hb!]
    \centering
    \caption{Physical and Software Parameters of the Platform}
    \resizebox{0.7\linewidth}{!}{%
     \begin{tabular}{ccc}
            \toprule
            \textbf{Group} &  \textbf{Parameter} &  \textbf{Value}\\
            \midrule
            \multirow{6}{*}{\makecell[c]{Module}}
                &  Contour Diameter $/m$ & $0.100$\\
                &  Height   $/m$ & $0.100$\\
                &  Weight   $/kg$ & $0.253$\\
                &  omni-wheel Radius$/m$ & $0.028$\\
                &  Maximum Velocity$/m\cdot s^{-1}$ & $0.073$\\
                &  Maximum Payload $/kg$ & $0.330$\\
            \midrule
            \multirow{3}{*}{\makecell[c]{Docking \\Mechanism}}
                & $F_{\textit{tensile}}/N$   &$1.29$\\ 
                & $\tau_{\textit{Shear}}/N \cdot mm$   &$8.39$\\ 
                & Align range/$mm$   &$10$\\
            \midrule
            \multirow{2}{*}{\makecell[c]{Structure}}  
                & Communication delay/$ms$ & $20$ \\
                & Motor control rate/$Hz$ & $500$\\
            \bottomrule
        \end{tabular}
        }
    \label{tab:setup}
\end{table}

The PCB contains a wireless integrated MCU (ESP32), by which the robots connect to a WiFi network, receive commands from the host computer, and control the motors. By the 2DoF steering and driving mechanism, a single robot can navigate to the target position, dock to/undock from the current formation, and change the docking position continuously. Meanwhile, the steering direction of each robot can be modified without breaking current formation, and the swarm can move exactly like a single omnidirectional robot. The important physical and software parameters are summarized in \cref{tab:setup}.

\begin{figure}[t!]
	\centering
    \includegraphics[width=\linewidth]{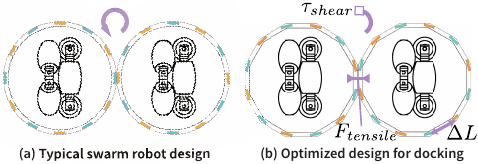}
    \caption{\textbf{Docking mechanism design.} (a) Magnetic docking mechanism applied to a round robot contour could easily slip and break the formed structure. (b) Our polygon-like contour design provides extra shear strength for robots to better maintain formation. }
	\label{fig:docking}
 \end{figure}

\subsection{Docking Mechanism}
\label{sec:dock_design}
We develop an efficient and flexible docking mechanism by utilizing a permanent magnet array, as is shown in \cref{fig:docking}. Magnets are arranged around the contour of the robot, and  adjacent magnets are always in an opposite direction. When robots get close to each other, the magnets array can passively align and attract, making robots docked together. 

\cref{fig:docking}a illustrates the basic setup of the magnets array with a round robot contour. The robot can easily transfer to a nearby docking position by moving itself. However, the transfer requires very slight shear torque, making the formation sensitive to disturbance. As a result, we optimize the robot contour as is shown in \cref{fig:docking}b. Compared to the round contour, the polygon-like contour can provide more shear strength when transferring between docking positions. And the polygon-like contour can also provide more contact area and generate better stability when transferring objects. Under the same placement of magnets, extra shear torque  $\tau_{\textit{Shear}} = F_{\textit{tensile}}\cdot\Delta L/2$ is required, in which $\Delta L$ differs according to the polygon used. In this way, we can adjust the balance between consistency and shear strength of docking. 

Considering the friction provided by the wheel, we utilize  $10mm\times5mm\times1mm$ NdFeB magnets for the array. 24 magnets are equally placed around a regular 24-gon contour of radius $5mm$. The magnets are installed $1mm$ inside the contour presenting a tensile strength of $1.29N$ between robots, and the shear strength is approximately calculated by $\tau_{\textit{Shear}} = F_{\textit{tensile}}\cdot\Delta L/2=1.29N\cdot 6.5mm = 8.39\times10^{-3}N\cdot m$. 

\begin{figure}[t!]
    \centering  
    \begin{subfigure}[b]{\linewidth}
    \centering\includegraphics[width=0.85\linewidth]{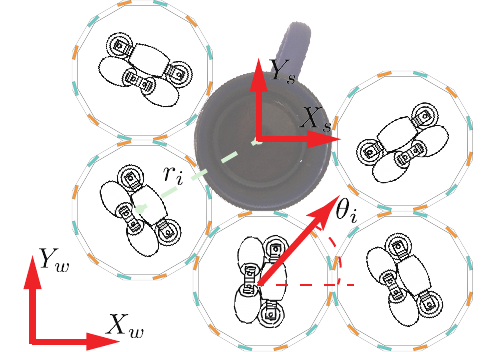}
    \end{subfigure}%
    \caption{\textbf{The coordinate system of the structure.} The origin of the structure coordinate locates at its geometric center, and the heading direction of each module $\theta_i$ refers to the angle between the direction of the omni-wheel and the structure coordinate's $X$ positive direction.}
    \label{fig:kinematics}
\end{figure}

\section{System Model \& Definitions} \label{sec:model}
\textbf{Definition 1 (Module)} \textit{refers to as the \ac{wmr} can dock with other modules in the horizontal plane.}

\textbf{Definition 2 (Structure)} \textit{refers to as the \ac{owmr} composed by $n$ docked modules ($n\geq3$).}

\subsection{Module Kinematics}
\label{sec:single}
The kinematics of a single module can be described as~\cite{song2004design}:
\begin{equation}
\small
\begin{aligned}
    v_{ix}&=v_{i}\cos\theta_i=\omega_{i}R\cos\theta_i,\\
    v_{iy}&=v_{i}\sin\theta_i=\omega_{i}R\sin\theta_i,
\end{aligned}
\end{equation}
where $\theta_i$ is the direction of the \ac{wmr}, $v_{ix}$ and $v_{iy}$ are the x and y components of its velocity $v_i$, $R$ is the radius of the omni-wheel.
Of note, with our customized design, each module is dynamically similar to the differential drive \ac{wmr}, and it receives input $\theta_i$ for heading direction control and $\omega_i$ for omni-wheel velocity control.

\begin{figure*}[ht!]
    \centering  
    \includegraphics[width=0.9\linewidth,trim=0cm 4cm 0cm 1.5cm, clip]{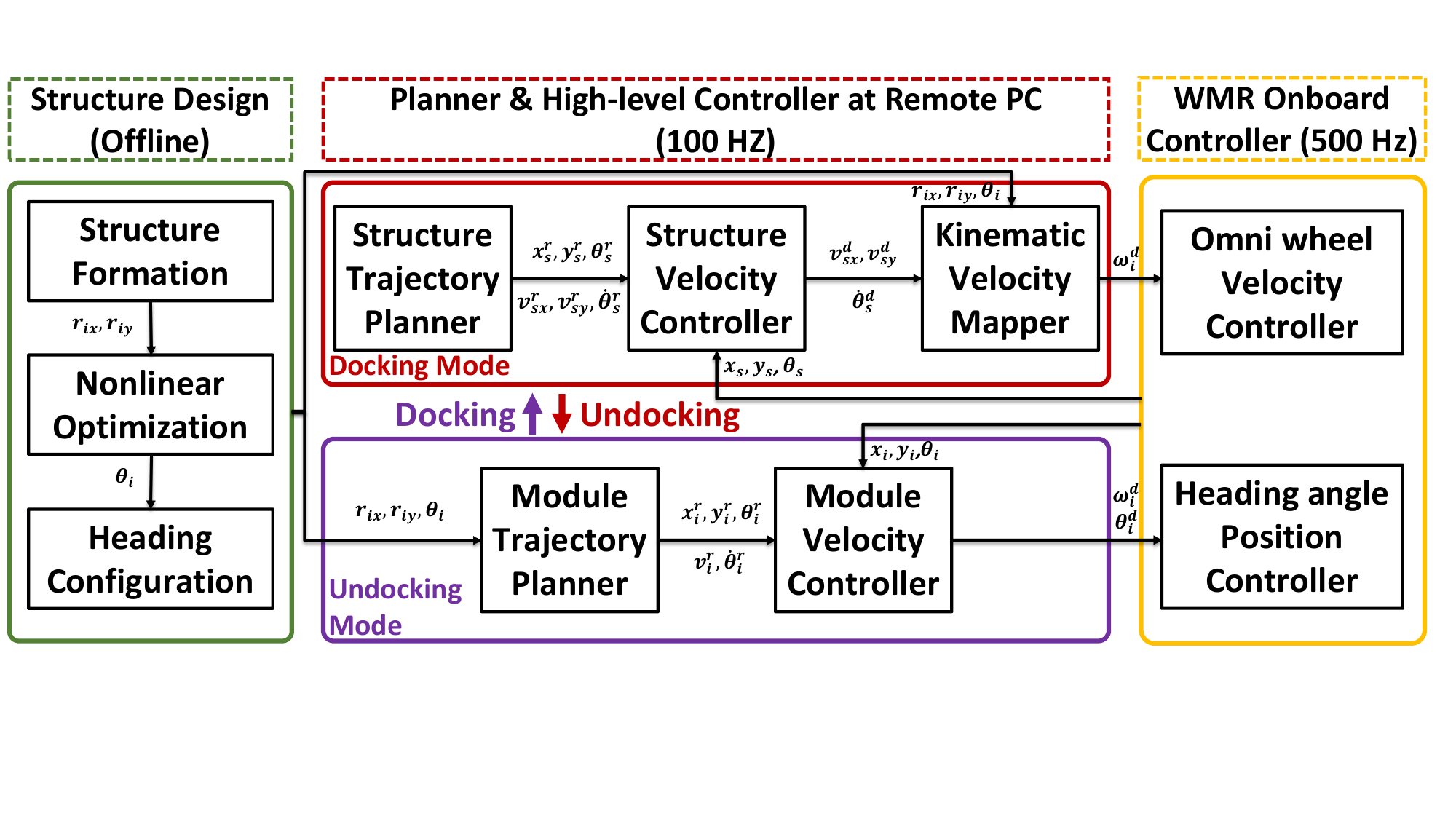}
    \caption{\textbf{Hierarchical Control architecture.} (i) Given a designed structure, the optimal heading configuration is computed through nonlinear optimization. (ii) Each module picks its target position and heading from the structure and tracks the generated trajectory with the module controller before docking. (iii) Once a structure is formed after docking, it is treated as an \ac{owmr}. the structure controller would calculate the desired velocity command for trajectory tracking and map that to the desired angular velocity of each wheel.}
    \label{fig:control}
\end{figure*}

\subsection{Structure Kinematics}
With $n$ rigidly connected \ac{wmr} modules, the connected structure can be treated as one omnidirectional \ac{wmr}. As shown in \cref{fig:kinematics}, choosing the geometric center of the structure as the origin (structure frame $\{X_s,Y_s\}$), the velocity of $i$-th \ac{wmr} module $v_i$ can be written as:
\begin{equation}
\small
    v_i=(v_{sx}-\dot{\theta}_sr_{i,y})\cos\theta_i+(v_{sy}+\dot{\theta}_sr_{i,x})\sin\theta_i,
\end{equation}
where $v_{sx}$ and $v_{sy}$ represent the structure's linear velocities along the x and y axis in world frame, $\dot{\theta}_s$ is the angular velocity, $r_{i,x}$ and $r_{i,y}$ are the position vector's x and y components from the structure origin to $i$-th \ac{wmr}'s center. Combining with the kinematics equations analyzed of each module and turning the results into matrix form we can get~\cite{almasri2021modeling}:
\begin{equation}
\small
    \pmb{\Omega}=\pmb{M}\pmb{V}
    \label{eq:kinematics}
\end{equation}
where
\begin{equation}
\small
    \pmb{\Omega}=
    \begin{bmatrix}
    \omega_1 \\
    \vdots\\
    \omega_{n}
    \end{bmatrix}, 
    \pmb{V}=
    \begin{bmatrix}
    v_{sx}      \\
    v_{sy} \\
    \dot{\theta}_s
    \end{bmatrix},
\end{equation}
and
\begin{equation}
\small
    \pmb{M}=\frac{1}{R}
    \begin{bmatrix}
   \cos\theta_1 & \sin\theta_1 & r_{1,x}\sin\theta_1-r_{1,y}\cos\theta_1 \\
    \vdots      & \vdots & \vdots \\
   \cos\theta_n & \sin\theta_n & r_{n,x}\sin\theta_n-r_{n,y}\cos\theta_n \\
    \end{bmatrix}.
\end{equation}

\textbf{Definition 3 (Heading Configuration)} \textit {refers to as the heading angle $\theta_i$ of each module w.r,t the structure frame.}

\textbf{Definition 4 (Formation Configuration)} \textit {refers to as the position vector $r_{i,x}$, $r_{i,y}$ of each module and the topology information that describes the docking faces between modules. }
\section{Control}\label{sec:control}
The overall controller has a hierarchical architecture, as shown in \cref{fig:control}.

\subsection{Module Control}
The desired linear $v_i^d$ and angular velocity $\dot{\theta}_i^d$ of the single module is designed as: 
\begin{equation}
\small
    \begin{bmatrix}
        v_i^d\\ \dot{\theta}_i^d
    \end{bmatrix}
    =
    \begin{bmatrix}
        {v_i^r}{\cos}e_{s\theta}+k_{s1}e_{ix}
        \\ \dot{\theta}_i^r+k_{s2}v_i^re_{iy}+k_{s3}v_i^r{\sin}e_{i\theta}
    \end{bmatrix},
\end{equation}
where $v_i^r$ and $\dot{\theta}_i^r$ are reference linear and angular velocity, $k_{s1}$, $k_{s3}$, and $k_{s3}$ are positive parameters, and $e_{ix}$, $e_{iy}$, and $e_{i\theta}$ are
errors between the reference trajectory and the real trajectory
in $x$, $y$, and $\theta$ directions respectively,
\begin{equation}
\small
     e_{ix} = x_i^r - x_i, e_{iy} = y_i^r - y_i, e_{i\theta} = \theta_i^r - \theta_i .
\end{equation}%
The desired heading direction is calculated with discrete integration: 
\begin{equation}
\small
    \theta_i^d=\theta_i+\dot{\theta}_i^d\,dt.
\end{equation}

\subsection{Heading Configuration Optimization}
\label{sec:headingopt}
Assuming the formation configuration of the structure is given, we formulate a nonlinear optimization problem to find the optimal heading configuration for the structure. According to the kinematics relationship \cref{eq:kinematics},
we can calculate the energy consumption of the whole structure as:
\begin{equation}
\small
    \sum_{i=1}^n {\omega_i^2}=\pmb{\Omega}^T\pmb{\Omega}=\pmb{V}^T\pmb{M}^T\pmb{M}\pmb{V}\leq({\sigma}_\textit{max}(\pmb{M})\lVert\pmb{V}\rVert_2)^2,
    \label{eq:energy}
\end{equation}
where ${\sigma}_\textit{max}(\cdot)$ is he maximum singular value of a matrix. On the other hand, the omni-direction goal
requires $\pmb{M}$ in full row rank. Therefore, the objective function is designed as:  
\begin{equation}
\small
    \operatorname*{argmin}_{\theta_{1},\cdots,\theta_{n}}  \quad
    \textit{cond}(\pmb{M})+{\sigma}_\textit{max}(\pmb{M})^2,
\label{eq:fitness}
\end{equation}
where $\textit{cond}(\cdot)$ is the condition number of a matrix, respectively. The left half $\textit{cond}(\pmb{M})$ considers the controllability of the system and ensures the connected structure is omnidirectional ($\textit{cond}(\pmb{M})=\textit{Inf}$ if $\textit{rank}(\pmb{M})<3$); while the right half ${\sigma}_\textit{max}(\pmb{M})^2$ characterize energy consumption minimization \cref{eq:energy}. The inequality constraints are designed as:  
\begin{equation}
\small
    0\leq\theta_i\leq2\pi, \quad\quad  \forall i=1,\cdots,n
\end{equation}

To build the target structure with modules, the optimized heading configuration and the structure formation will be utilized by the trajectory planner of each module as the final configuration to plan the docking motion. After this docking process, the heading angle $\theta_i$ of each module is fixed and the structure is controlled as an omnidirectional \ac{wmr} with constant velocity mapper ($\pmb{M}$ in \cref{eq:kinematics}).

\subsection{Structure Control}
\label{sec:s_control}
With the optimal structure configuration, the tracking controller of the \ac{owmr} structure is designed as~\cite{shi2020path}: 
\begin{equation}
\small
    \begin{bmatrix}
    v_{sx}^d      \\
    v_{sy}^d \\
    \dot{\theta}_s^d
    \end{bmatrix}=
    \begin{bmatrix}
    v_{sx}^r+k_{{x1}}e_{sx}+k_{{x}2}\int{e_{sx}}\,dt\\
    v_{sy}^r+k_{{y1}}e_{sy}+k_{{y}2}\int{e_{sy}}\,dt\\
    \dot{\theta}_s^r+k_{\theta1}e_{s\theta}+k_{\theta2}\int{e_{s\theta}}\,dt\,
    \end{bmatrix},  
\end{equation}
where $v_{sx}^r$, $v_{sy}^r$, $\dot{\theta}_s^r$ are reference velocities in three \ac{dof}, $k_{xi}$, $k_{yi}$, and $k_{\theta{i}}$ are positive control gains, 
$e_{sx}$, $e_{sy}$, and $e_{s\theta}$ are the tracking errors. 
Then the desired velocities $\pmb{\Omega}$ of the omni-wheels can be calculated with \cref{eq:kinematics}.

\section{Simulation}\label{sec:sim}
In the simulation, we evaluate whether the optimal heading configuration computed by our optimization frame is indeed the most efficient in generating omnidirectional movements. 

\begin{figure}[t!]
 \centering  
    \begin{subfigure}[b]{0.95\linewidth}
    \centering
    \includegraphics[width=\linewidth]{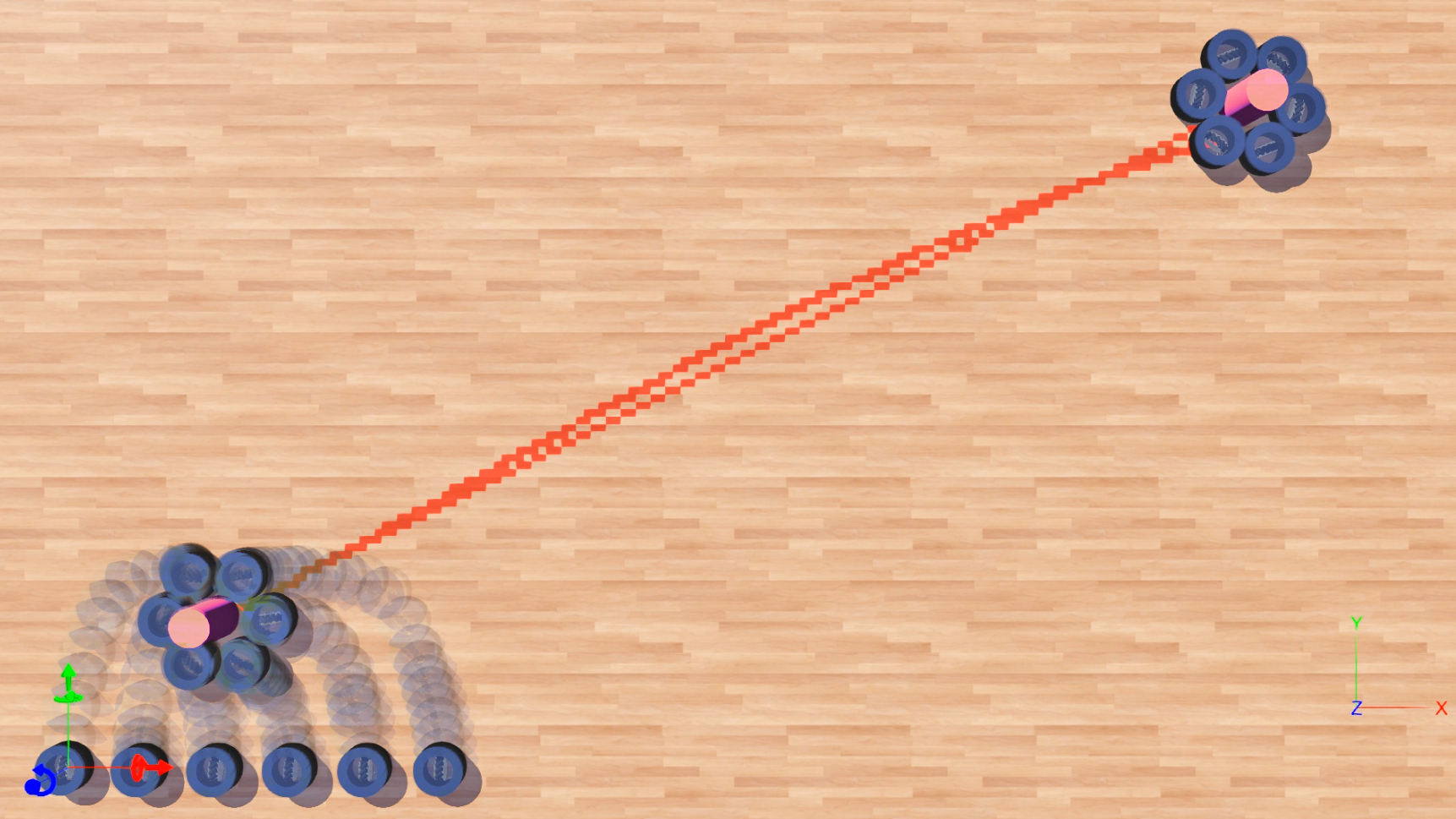}
    \caption{Visualization of the simulated object transportation task}
    \label{fig:trans_shot}
    \end{subfigure}
    \\
    \begin{subfigure}[b]{0.95\linewidth}
    \centering
    \includegraphics[width=\linewidth]{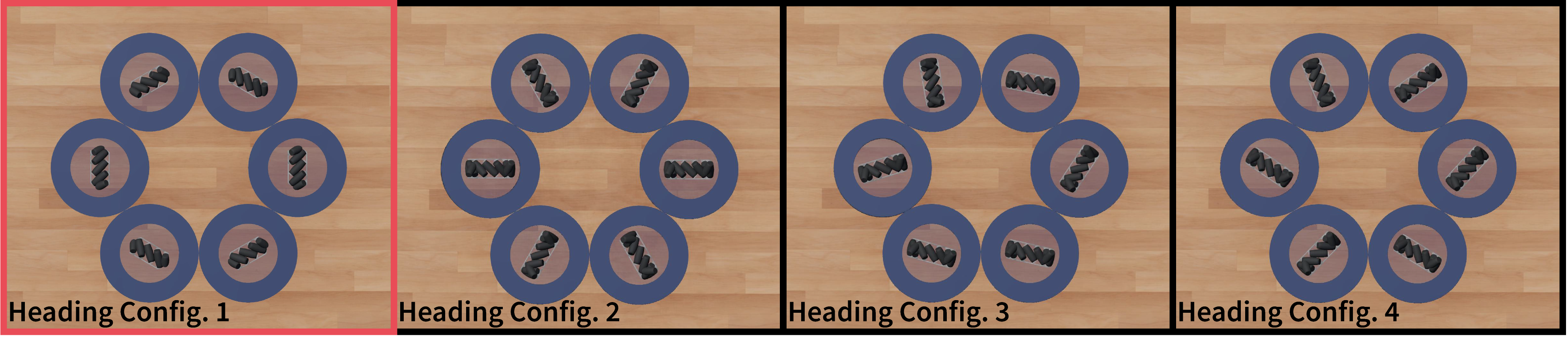}
    \caption{Four heading configurations for evaluation}
    \label{fig:config}
    \end{subfigure}%
    \\
    \begin{subfigure}[b]{\linewidth}
    \centering
    \includegraphics[width=\linewidth]{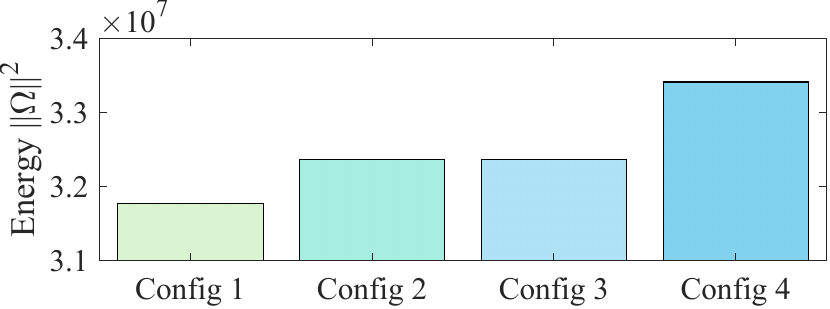}
    \caption{The energy cost.}
    \label{fig:trans_pos}
    \end{subfigure}%
    \caption{\textbf{Simulation study.} In (a) the transportation process can be divided into two stages: first build the target structure around the object, and then transport the object along the trajectory as an \ac{owmr}. In (b), four heading configurations are compared in the same transportation task with their consumed energy plotted in (c) where \textit{Config. 1} solved by our optimization provides the best energy efficiency.}   
    \label{fig:transport}
\end{figure}

\textbf{Simulation Setup:} We use WeBots\cite{Webots04} as the physical simulator to measure the efficiency of the structure formed by six robots during collaborative object transportation. A PROTO is designed to describe the structure and properties of the robot with parameters listed in \cref{tab:setup}. All robots utilize the same on-board controller for motor position and velocity control which takes commands from a high-level controller introduced in \cref{sec:control}. The simulation environment is shown in \cref{fig:trans_shot}.

\textbf{Simulation Result:} \cref{fig:config} lists four heading configurations we compare in the study; the \textit{Config. 1} in the red box is the optimal one computed, whereas the other three are manually designed. We use the sum of the square of each wheel's angular velocity as the indicator of the energy efficiency of the structure (\cref{eq:energy}).
Despite the team can successfully transport the object to the target with all four heading configurations with omnidirectional movements, the one computed by our optimization framework consumes the least energy; see \cref{fig:trans_pos}. The result indicates the necessity of adjusting heading configuration to achieve more efficient locomotion during collaborative object transportation.

\section{Experiment}
\label{sec:exp}
\subsection{Experiment Setup}
To further demonstrate the capability introduced by our platform, we experiment with the proposed modular robot system in the physical world. Specifically, we use the Vicon motion capture system (MoCap) to measure the position and heading of each module. The trajectory planner and high-level controller of the system run on a remote PC, which communicates with the MoCap through Ethernet with 100~$Hz$. The high-level controller calculates the desired angular velocity $\omega_i^d$ and heading angle $\theta_i^d$ for each module and sends them through WiFi. Each module is embedded with an onboard omni-wheel velocity controller and heading angle position controller.

Four cases of experiments are conducted. we first test the trajectory tracking performance of a single module (\textit{Case~1}) and an omnidirectional structure with six modules (\textit{Case~2}). Then, in \textit{Case~3}, we present the whole collaborative object transportation process with six modules docked to cage the object. Finally, we study the object transportation performance of a connected structure with different payloads in \textit{Case~4}. 

\subsection{Experiment Results}
\begin{figure}[ht!]
    \centering  
    \includegraphics[width=\linewidth,trim=0cm 0cm 0cm 0cm,clip]{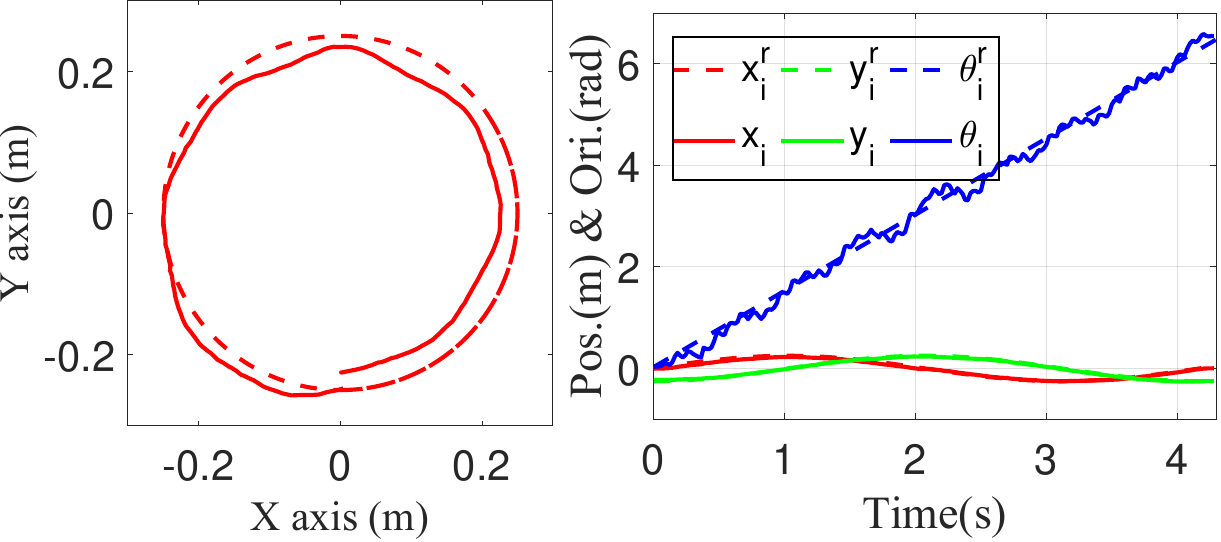}
    \caption{\textbf{Case 1: Single Module.} The robot with 2 controllable \ac{dof} is capable of tracking a circular trajectory which satisfies the nonholonomic constraint.}
    \label{fig:exp_single}
\end{figure}

\textbf{Case 1: Single Module.} A circular trajectory with $R=0.25~m$ is utilized as the reference to test the moving capability of the single module. Both the reference trajectory (dashed line) and the tracking trajectory (solid line) are plotted on the left of \cref{fig:exp_single}, while the detailed tracking performance of each \ac{dof} is plotted on the right. Given the negative effects brought by the asymmetric friction of the omni-wheel, the robot is still capable of performing good navigation.  

\begin{figure}[t!]
    \centering  
    \begin{subfigure}[b]{0.47\linewidth}
    \centering
    \includegraphics[width=\linewidth, trim=2cm 1cm 1cm 2cm,clip]{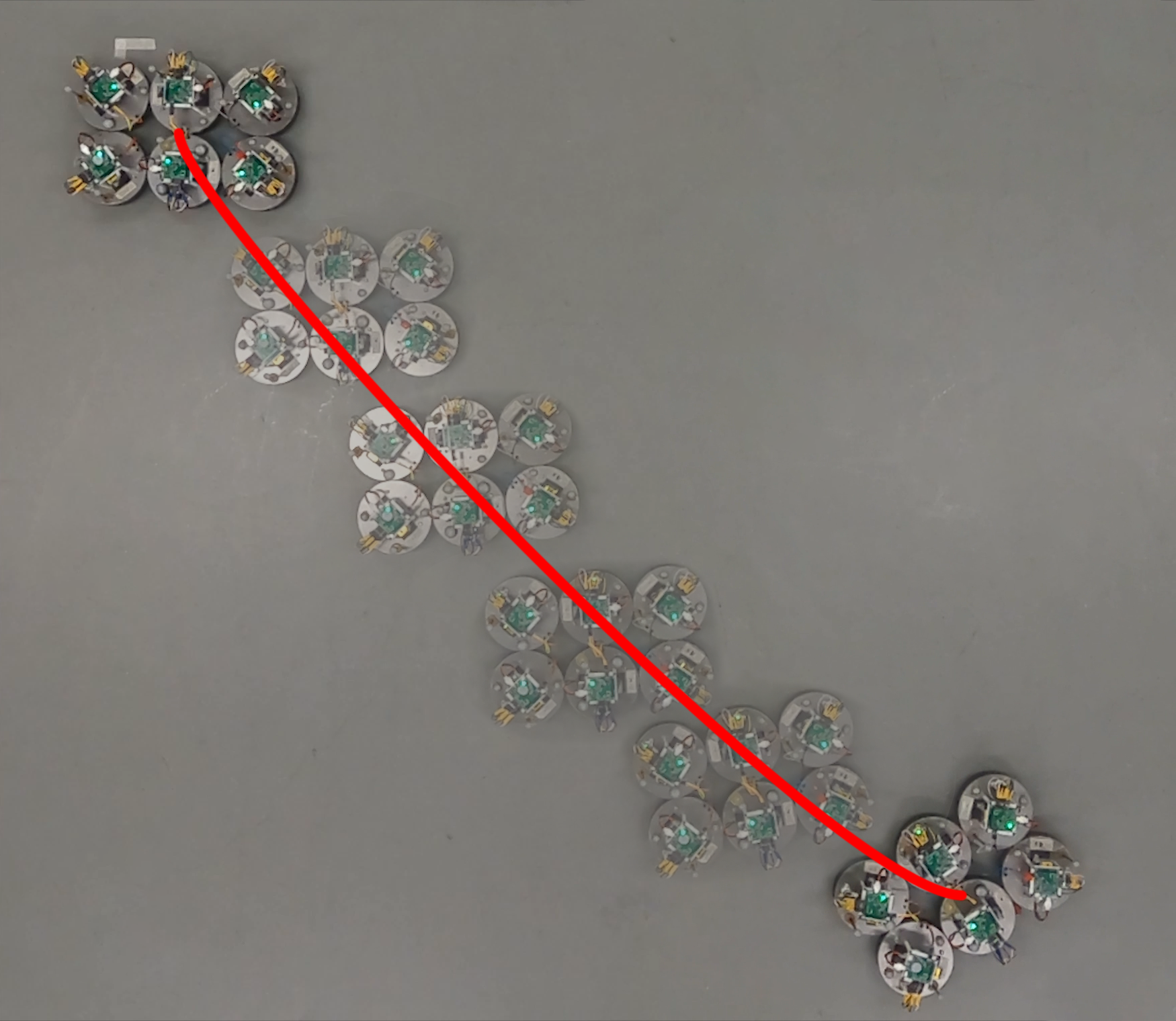}
    \end{subfigure}%
    \begin{subfigure}[b]{0.53\linewidth}
    \centering
    \includegraphics[width=\linewidth, trim=0cm 0cm 0cm 0cm, clip]{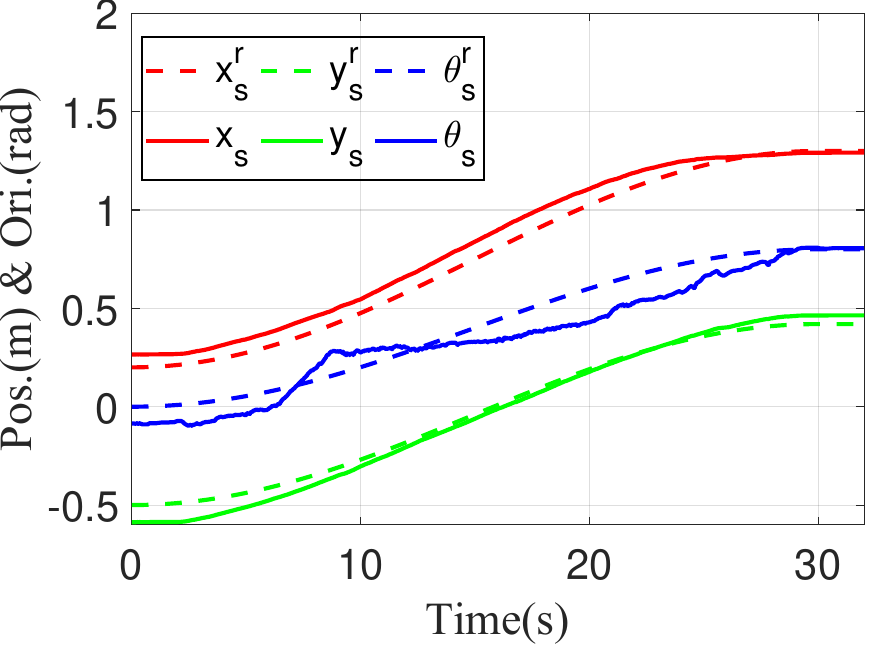}
    \end{subfigure}%
    \caption{\textbf{Case 2: Omnidirectional Structure.}  The connected structure with six modules is over-actuated and can be controlled as an \ac{owmr} to track a 3-DOF trajectory stably. } 
    \label{fig:exp_multi}
\end{figure}

\textbf{Case 2: Omnidirectional Structure.} A rectangular structure composed of six modules is built, where the heading angle of each module is optimized with the formulation introduced in \cref{sec:headingopt}. As shown in \cref{fig:exp_multi}, it can maintain the formation stably with magnetic force and track a 3-\ac{dof} reference trajectory accurately.

\begin{figure}[t!]
    \centering  
    \begin{subfigure}[b]{\linewidth}
    \centering
    \includegraphics[width=\linewidth,trim=8cm 0cm 2cm 10cm, clip]{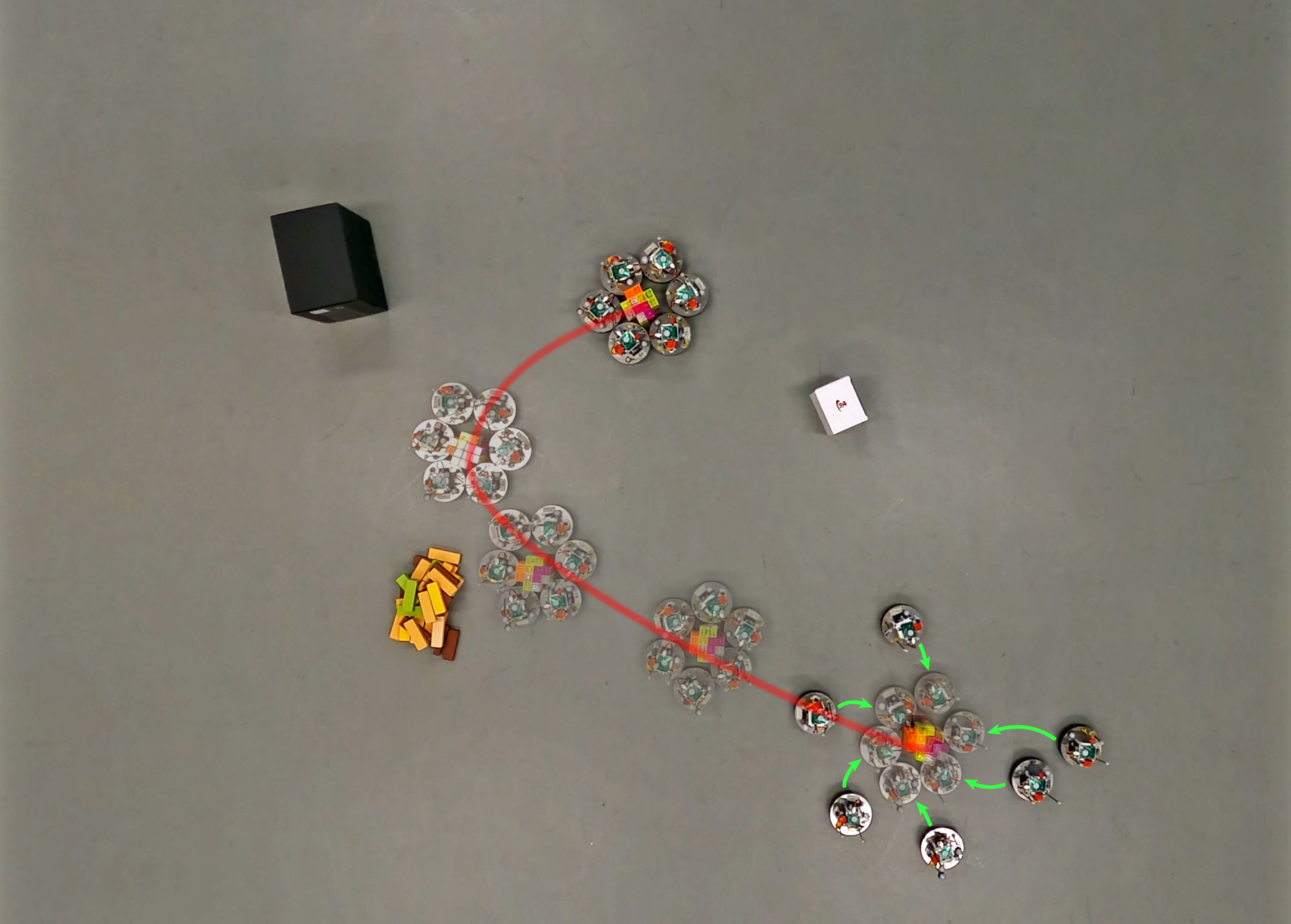}
    \caption{Aggregating and transporting process.}
    \label{fig:exp_trans_shot}
    \end{subfigure}\\
    \begin{subfigure}[b]{\linewidth}
    \centering
    \includegraphics[width=\linewidth,trim=0cm 0cm 0cm 0cm,clip]{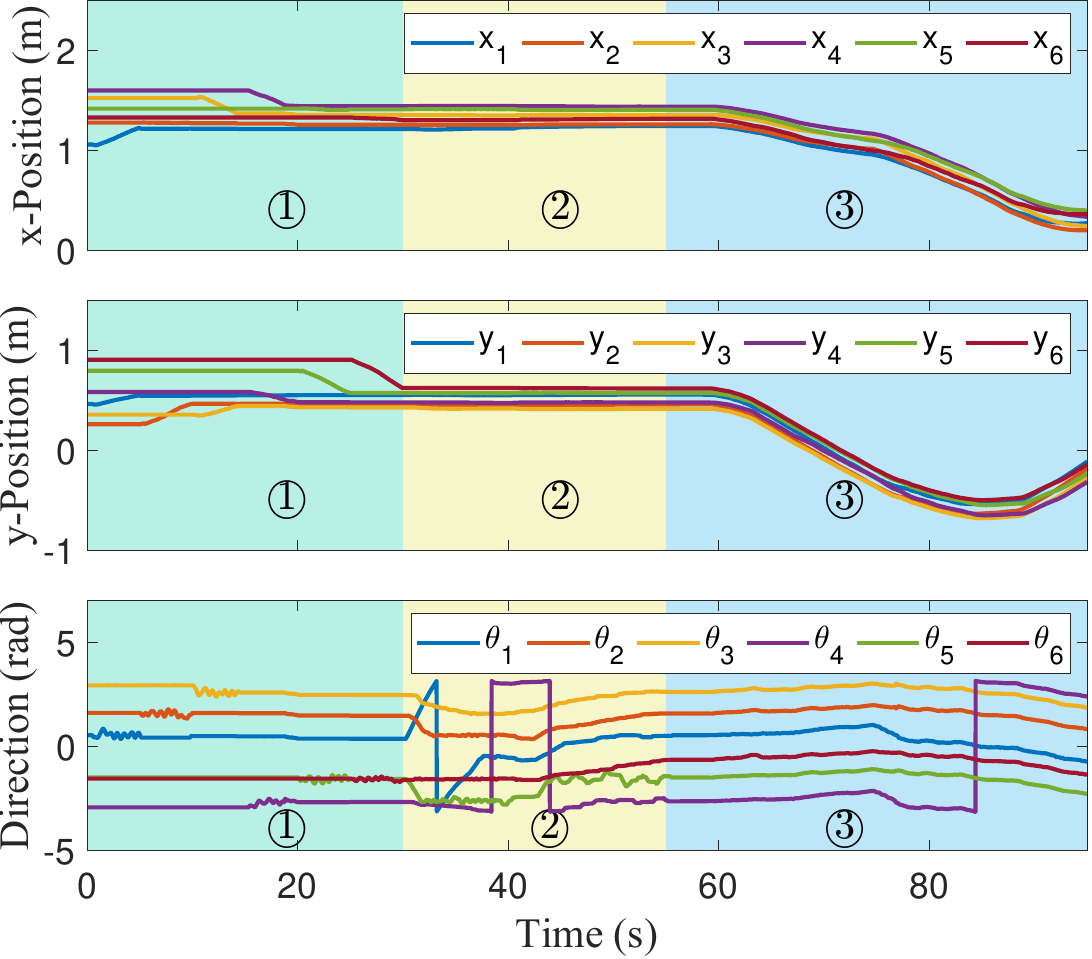}
    \caption{Position and direction of each module.}
    \label{fig:exp_trans_pos}
    \end{subfigure}%
    \caption{\textbf{Case 3: Object Transportation.} (a) shows the process consisting of aggregating (green) and transporting (red). (b) illustrates the change of position and direction of each module. }  
    \label{fig:exp_transport}
\end{figure}
\textbf{Case 3: Object Transportation.} A hexagon-shaped structure for six robots is designed to cage a cylindrical object and collaboratively transport it to a target position (see \cref{fig:exp_trans_shot}). The task can be divided into three stages: in stage \textcircled{1}, each module is required to move to the target docking position from the initial position; in stage \textcircled{2} each module adjusts its heading and position to form the structure around the object; in stage \textcircled{3}, each module first reorients its wheel's heading direction to the optimized one and then drives along the trajectory to transport the object passively. The position and direction of each member are plotted in \cref{fig:exp_trans_pos} in detail.

\begin{figure}[t!]
    \centering  
    \begin{subfigure}[b]{0.5\linewidth}
    \centering
    \includegraphics[height=0.9\linewidth, ]{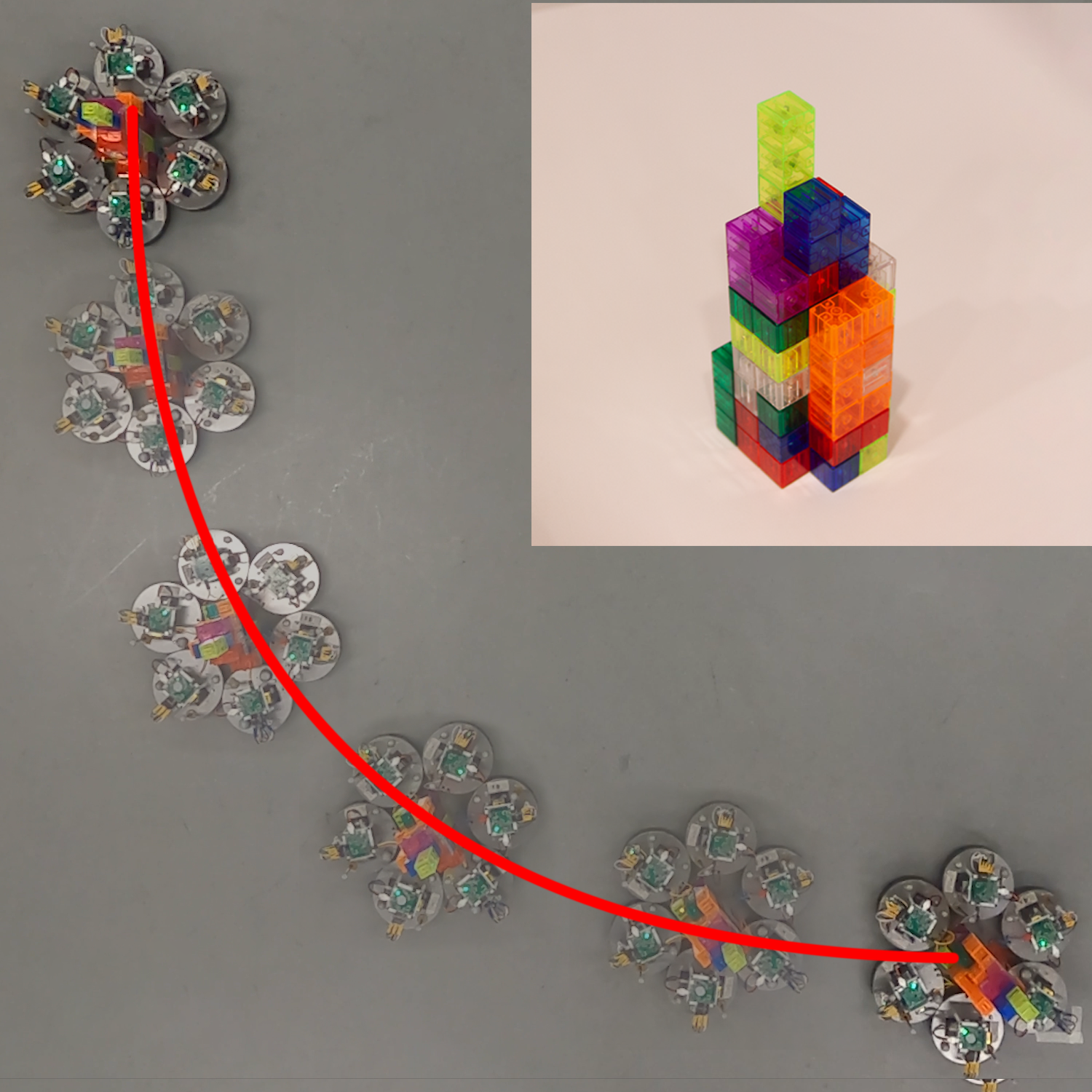}
    \end{subfigure}%
    \begin{subfigure}[b]{0.5\linewidth}
    \centering
    \includegraphics[height=0.9\linewidth, ]{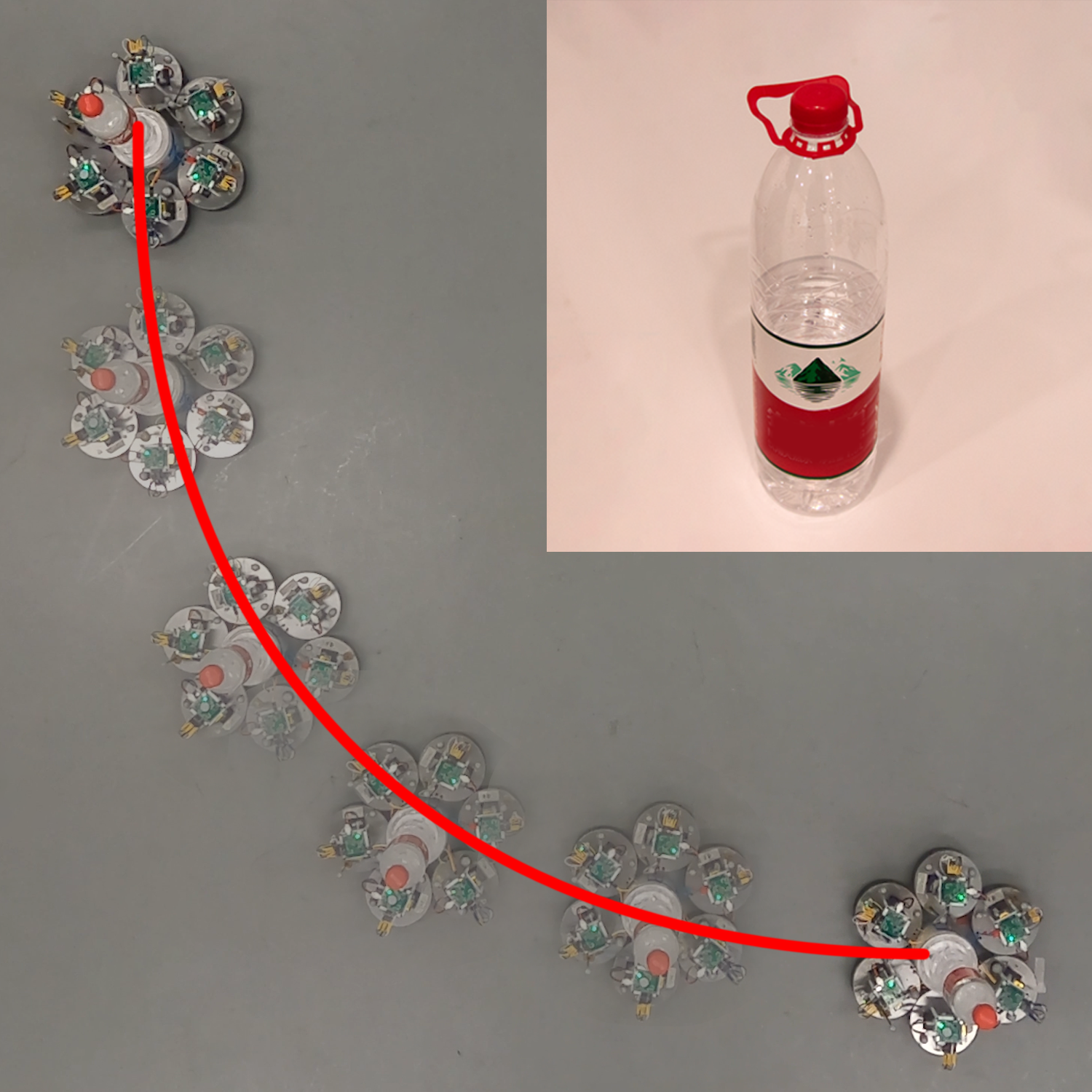}
    \end{subfigure}\\
    \begin{subfigure}[b]{0.5\linewidth}
    \centering
    \includegraphics[width=\linewidth,trim=0cm 0cm 0cm 0cm, clip]{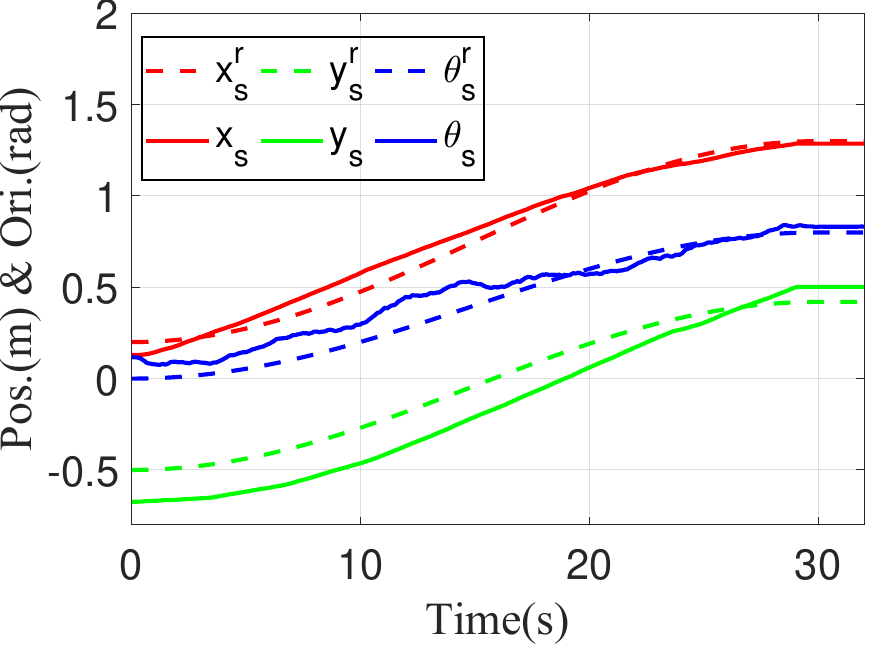}
    \caption{$0.3~kg$ payload}
    \label{fig:low_payload}
    \end{subfigure}%
    \begin{subfigure}[b]{0.5\linewidth}
    \centering
    \includegraphics[width=\linewidth, trim=0cm 0cm 0cm 0cm, clip]{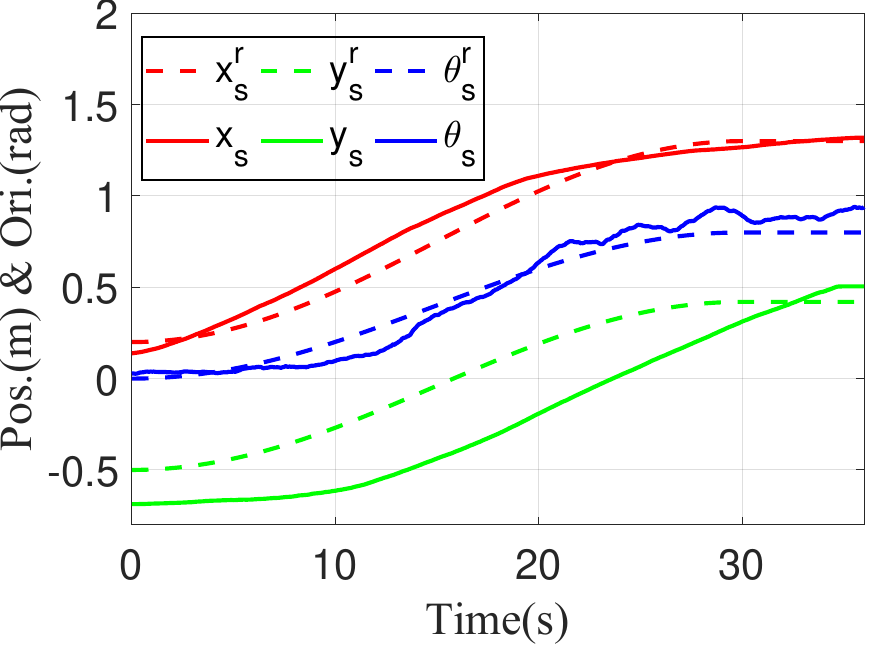}
    \caption{$1.2~kg$ payload}
    \label{fig:high_payload}
    \end{subfigure}%
    \caption{\textbf{Case 4: Payload Carrying.} While the tracking performance degrades as the payload increases, the team managed to navigate to the target point.}
    \label{fig:exp_multi_payload}
\end{figure}

\textbf{Case 4: Payload Carrying.} Through the same hexagon-shaped structure, six robots are tasked to transport objects with different weights along a trajectory with large curvature to verify the robustness of the formation and the controller. As shown in \cref{fig:exp_transport}, two objects with a weight of $0.3~kg$ and $1.2~kg$ successfully transported by the structure along the reference trajectory. By caging the objects tightly, the orientation of the objects can also be accurately controlled by the structure.
In \cref{fig:exp_multi_payload}, the tracking performance degrades as the payload increases, which is due to the larger friction force introduced by the object. This issue can be improved by utilizing model-reference adaptive controller~\cite{su2022object} to replace the structure tracking controller introduced in \cref{sec:s_control}

\section{Conclusion} \label{sec:con}
In this paper, we presented a neoteric solution to the collaborative object transportation problem using a self-reconfigurable robot team with the capability of omnidirectional movements. We designed and built a 2-DoF WMR with an omni-wheel, an optimized contour, and a dockable magnet array. A decoupled planning and control framework was also propounded to enable our robot team to perform the entire pipeline of collaborative transportation, including heading configuration optimization, self-navigation, docking, and collaborative motion. We also demonstrated the effectiveness of our solution in both a simulation and a real-world experiment.

The proposed robot system offers a promising approach to solving complex transportation tasks. By utilizing magnetic docking and self-reconfiguration capabilities, the proposed system can efficiently move objects in any direction while maintaining their relative positions. This feature makes it ideal for scenarios where precise object positioning is crucial, such as in manufacturing, warehousing, and logistics operations. Additionally, the modular design of the proposed robot allows for scalability and flexibility, making it suitable for a wide range of tasks, from small-scale object transportation to large-scale material handling. Furthermore, the use of optimization-based methods and hierarchical controllers ensures that the system can adapt to changes in the environment and operate in a stable and efficient manner.

\setstretch{0.96}
\paragraph*{Acknowledgement}
The authors would like to thank Chi Chu at BIGAI and Kuntong Han at THU  for their help with experiments.

{

\small
\bibliographystyle{ieeetr}
\bibliography{reference}
}
\end{document}